\documentclass[pdflatex,sn-mathphys-num]{sn-jnl}% Math and Physical Sciences Numbered Reference Style
\usepackage{graphicx}%
\usepackage{multirow}%
\usepackage{amsmath,amssymb,amsfonts}%
\usepackage{amsthm}%
\usepackage{mathrsfs}%
\usepackage[title]{appendix}%
\usepackage{xcolor}%
\usepackage{textcomp}%
\usepackage{manyfoot}%
\usepackage{booktabs}%
\usepackage{algorithm}%
\usepackage{algorithmicx}%
\usepackage{algpseudocode}%
\usepackage{listings}%
\usepackage{ulem}%

\theoremstyle{thmstyleone}%
\theoremstyle{thmstyletwo}%

\theoremstyle{thmstylethree}%

\begin{document}

\title[Article Title]{R-GenIMA: Integrating Neuroimaging and Genetics with Interpretable Multimodal AI for Alzheimer’s Disease Progression}

% Alex Leow
\author[1]{\fnm{Kun} \sur{Zhao}}\email{kun.zhao@pitt.edu}

\author[1]{\fnm{Siyuan} \sur{Dai}}\email{siyuan.dai@pitt.edu}

\author[2]{\fnm{Yingying} \sur{Zhang}}\email{yingying.zhang01@utrgv.edu}

\author[3]{\fnm{Guodong} \sur{Liu}}\email{Guodong.liu@lilly.com}

\author[2]{\fnm{Pengfei} \sur{Gu}}\email{pengfei.gu01@utrgv.edu}

% \author[2]{\fnm{Erik} \sur{Enriquez}}\email{erik.enriquez01@utrgv.edu}

% \author[2]{\fnm{Dongchul} \sur{Kim}}\email{dongchul.kim@utrgv.edu}

\author[4]{\fnm{Chenghua} \sur{Lin}}\email{chenghua.lin@manchester.ac.uk}

\author[5]{\fnm{Paul M.} \sur{Thompson}}\email{pthomp@usc.edu}

\author[6]{\fnm{Alex} \sur{Leow}}\email{alexfeuillet@gmail.com}

\author[7]{\fnm{Heng} \sur{Huang}}\email{heng@umd.edu}

\author[8]{\fnm{Lifang} \sur{He}}\email{lih319@lehigh.edu}

\author*[1]{\fnm{Liang} \sur{Zhan}}\email{liang.zhan@pitt.edu}

\author*[2]{\fnm{Haoteng} \sur{Tang}}\email{haoteng.tang@utrgv.edu}

\author[9]{for the Alzheimer’s Disease Neuroimaging Initiative (ADNI) Project}

\affil*[1]{\orgdiv{Electrical \& Computer Engineering}, \orgname{University of Pittsburgh}, \orgaddress{\street{4200 Fifth Avenue}, \city{Pittsburgh}, \postcode{15260}, \state{PA}, \country{USA}}}

\affil*[2]{\orgdiv{Computer Science}, \orgname{University of Texas Rio Grande Valley}, \orgaddress{\street{1201 West University Drive}, \city{Edinburg}, \postcode{78539}, \state{TX}, \country{USA}}}

\affil[3]{\orgname{Eli and Lilly company}, \orgaddress{\street{893 S Delaware St}, \city{Indianapolis}, \postcode{46285}, \state{IN}, \country{USA}}}

\affil[4]{\orgdiv{Computer Science}, \orgname{The University of Manchester}, \orgaddress{\street{Kilburn Building, Oxford Road}, \city{Manchester}, \postcode{M13 9PL}, \country{UK}}}

\affil[5]{\orgdiv{Imaging Genetics Center}, \orgname{University of Southern California}, \orgaddress{\street{4676 Admiralty Way}, \city{Marina del Rey}, \postcode{90292}, \state{CA}, \country{USA}}}

\affil[6]{\orgdiv{Psychiatry}, \orgname{University of Illinois Chicago}, \orgaddress{\street{1601 W. Taylor St.}, \city{Chicago}, \postcode{60612}, \state{IL}, \country{USA}}}

\affil[7]{\orgdiv{Computer Science}, \orgname{University of Maryland College Park}, \orgaddress{\street{8125 Paint Branch Drive}, \city{College Park}, \postcode{20742}, \state{MD}, \country{USA}}}

\affil[8]{\orgdiv{Computer Science \& Engineering}, \orgname{Lehigh University}, \orgaddress{\street{113 Research Drive}, \city{Bethlehem}, \postcode{18015}, \state{PA}, \country{USA}}}


\abstract{
Early detection of Alzheimer's disease (AD) requires models capable of integrating macro-scale neuroanatomical alterations with micro-scale genetic susceptibility, yet existing multimodal approaches struggle to align these heterogeneous signals. 
We introduce R-GenIMA, an interpretable multimodal large language model that couples a novel ROI-wise vision transformer with genetic prompting to jointly model structural MRI and single nucleotide polymorphisms (SNPs) variations. 
By representing each anatomically parcellated brain region as a visual token and encoding SNP profiles as structured text, the framework enables cross-modal attention that links regional atrophy patterns to underlying genetic factors. 
Applied to the ADNI cohort, R-GenIMA achieves state-of-the-art performance in four-way classification across normal cognition (NC), subjective memory concerns (SMC), mild cognitive impairment (MCI), and AD. 
Beyond predictive accuracy, the model yields biologically meaningful explanations by identifying stage-specific brain regions and gene signatures, as well as coherent ROI–Gene association patterns across the disease continuum. Attention-based attribution revealed genes consistently enriched for established GWAS-supported AD risk loci, including \textit{APOE, BIN1, CLU, and RBFOX1}. 
Stage-resolved neuroanatomical signatures identified shared vulnerability hubs across disease stages alongside stage-specific patterns: striatal involvement in subjective decline, frontotemporal engagement during prodromal impairment, and consolidated multimodal network disruption in AD. 
These results demonstrate that interpretable multimodal AI can synthesize imaging and genetics to reveal mechanistic insights, providing a foundation for clinically deployable tools that enable earlier risk stratification and inform precision therapeutic strategies in Alzheimer's disease.
}




\maketitle

\section{Introduction}\label{sec1}
% Alzheimer's Disease, Why we can and Why we need to do early detection on AD?
% Current neuroimaging studies on AD progression analysis, limitation that Genetic data may address
% Why we need to combine genetic SNPs + T1 to analyze AD progression with multimodal data? T1--structural, SNPs
% Why we need to use ROI patitioned images instead of whole images？ Why we need gene-ROI association?
% AI technical challenges in using imaging-genetics data --> VLM
% Brief our method and our findings. 

Neurodegenerative disorders such as Alzheimer’s disease (AD) pose an escalating challenge to global health, with prevalence rising rapidly alongside population aging and no curative therapies currently available \cite{WHO2025dementia,xiaopeng2025global}. 
Despite decades of research, the biological processes that initiate cognitive decline remain incompletely understood, particularly during the earliest preclinical phases of the disease.
AD unfolds along a prolonged continuum characterized by subtle yet systematic disturbances in brain structure, neuronal connectivity, and molecular homeostasis that emerge years before overt symptoms. 
Across this trajectory—from normal cognition (NC) through subjective memory concerns (SMC) and mild cognitive impairment (MCI)—neural systems undergo progressive reconfiguration driven by amyloid accumulation, tau propagation, synaptic dysfunction, and neuroinflammatory processes \cite{zhang2018risk}. 
Mounting evidence indicates that these early physiological and circuit-level disruptions can be detected using sensitive neuroimaging markers and computational models, offering a crucial opportunity to identify individuals at heightened risk before irreversible neurodegeneration occurs \cite{jack2016t}. % Add more of our papers in the revision stage! 
Consequently, characterizing the brain alterations that precede clinical dementia is essential not only for improving early diagnosis and risk stratification, but also for advancing mechanistic insight into AD progression and enabling timely, precision therapeutic interventions.

Structural MRI has been widely used to characterize AD–related neurodegeneration, revealing hallmark patterns of hippocampal atrophy, cortical thinning, and large-scale network disruption that track disease severity and clinical progression \cite{jack2018nia,weiner2013alzheimer}. 
As a macro-scale modality, T1-weighted imaging captures the anatomical consequences of accumulating pathology, yet these morphological alterations often emerge only after substantial molecular injury has occurred. 
In the earliest stages—particularly the subtle transition from normal cognition to subjective memory concerns—structural changes may be minimal, highly variable across individuals, and confounded by normative aging, limiting the sensitivity of MRI-only approaches for early detection \cite{sperling2011toward,jessen2014conceptual}. 
Conversely, genetic variation reflects micro-scale, lifelong biological susceptibility, with single nucleotide polymorphisms (SNPs) influencing amyloid processing, tau propagation, synaptic maintenance, lipid metabolism, and neuroinflammatory pathways that precede and ultimately shape regional brain vulnerability \cite{bateman2012clinical,kunkle2019genetic}. 
However, SNPs models may provide weak disease predictive power because molecular risk does not directly map to clinical symptoms or macro-scale atrophy patterns. 
These complementary limitations underscore a central challenge in AD research: macro-scale imaging captures downstream structural consequences, while micro-scale genetics encodes upstream drivers of vulnerability, yet neither modality alone is sufficient to fully capture early disease processes. 
Integrating T1 MRI with genetic SNPs therefore offers a crucial opportunity to bridge molecular susceptibility and structural expression, enhance early-stage detection, and reveal biologically grounded pathways through which genetic risk becomes instantiated in the human brain.
Despite growing interest in imaging–genetic studies of Alzheimer’s disease, several major challenges continue to limit progress toward biologically grounded and clinically actionable models. 
Structural MRI and genetic data operate at fundamentally different biological scales: MRI captures macro-scale anatomical consequences of neurodegeneration, whereas SNPs reflect micro-scale molecular susceptibility that precedes structural injury. 
Traditional statistical approaches—such as sparse regression \cite{bertsimas2020sparse,stein2010voxelwise,weiner2013alzheimer}, canonical correlation analysis \cite{weenink2003canonical,chi2013imaging,du2020detecting}, and GWAS-based association \cite{uffelmann2021genome,shen2010whole,shen2014genetic,lambert2013meta,thompson2014enigma,xu2020chimgen,fu2024cross,elliott2018genome}—are limited in their ability to model high-dimensional SNPs, nonlinear gene–brain interactions, and subtle cross-modal correspondences \cite{xin2022review}. 
As a result, they may yield suboptimal or unstable estimates of how genetic risk manifests in regional brain vulnerability. 
% \textcolor{red}{
% More recent deep learning and transformer-based multimodal methods \cite{li2025imaging,wang2022alzheimer,venugopalan2021multimodal} provide improved flexibility, yet they remain limited by separate modality-specific encoders and shallow fusion mechanisms that prevent early-layer cross-modal attention, thereby failing to learn a unified representation space where SNPs and MRI-derived regional features jointly contribute to modeling cross-scale biological interactions.
% % 我们用Vision-Language model比现在的用multimodal deep learning, 用单纯的MLP或transformer来embed SNP有哪些好处？
% These models typically rely on superficial feature concatenation or shallow attention mechanisms, struggle to capture long-range genomic dependencies, and provide limited biological interpretability, hindering the identification of disease-relevant brain regions, influential genetic variants, and the cross-scale interactions that link them.}

Recent deep learning and transformer-based multimodal architectures\cite{shen2019brain,li2025imaging,wang2022alzheimer,venugopalan2021multimodal} have substantially improved the capacity to integrate heterogeneous biomedical data.
However, traditional multimodal deep learning approaches employ independent encoders for MRI and genetic data (e.g., 3D CNNs for neuroimaging and MLPs for SNPs), followed by a simple late-fusion step in which high-level features are concatenated.
This late-fusion paradigm introduces a fundamental bottleneck: concatenation collapses the inherent spatial structure of neuroimaging representations, preventing the model from aligning localized anatomical alterations—such as hippocampal or temporoparietal degeneration—with their potential genetic determinants\cite{golovanevsky2022multimodal}.
As a result, the fused representation becomes biologically ambiguous, limiting interpretability and weakening the ability to infer fine-grained gene–brain associations.
Furthermore, most existing deep learning approaches address only coarse diagnostic categories, rather than modeling the continuous trajectory of early disease states in which genetic and structural signals evolve jointly \cite{li2025imaging,wang2022alzheimer,venugopalan2021multimodal}.
Collectively, these limitations highlight the need for a unified, scalable, and interpretable framework capable of bridging micro-scale molecular risk with macro-scale neuroanatomical expression across the AD trajectory.

Multimodal large language models (MLLMs) offer a promising direction for addressing these challenges by enabling joint representation learning across heterogeneous modalities—including structural MRI and SNP sequences—within a single pretrained, high-capacity architecture. 
Unlike conventional deep learning models, MLLMs treat imaging features and genetic variants as structured ``tokens," allowing the model to flexibly encode nonlinear cross-modal interactions and to capture long-range dependencies intrinsic to genomic data \cite{brown2020language,radford2021learning}.
Despite recent progress in medical MLLMs, current state-of-the-art systems (e.g., Med-LLaVA \cite{li2023llava}, RadFM \cite{wu2025towards}) remain fundamentally limited for neuroimaging–genetics integration because they are architecturally constrained to 2-dimensional visual backbones. 
Even when adapted to accommodate 3D volumetric inputs, these models typically rely on aggressive global pooling to control token complexity, which collapses spatial specificity and suppresses subtle regional variations \cite{xin2025med3dvlm}. 
As a result, fine-grained neuroanatomical signatures—especially those localized to disease-sensitive cortical or subcortical regions—are diluted, hindering any attempt to resolve biologically meaningful gene–brain associations. 
This limitation motivates the need for a more anatomically structured visual representation so that regional structural variations can be preserved and biologically meaningful gene–brain interactions can be captured by the MLLM.

To address these limitations, we introduce a new ROI-wise vision encoder—Region-wise Vision Transformer (RiT)—together with a multimodal framework, R-GenIMA (\uline{R}OI–\uline{Gen}ome fused \uline{I}nterpretable \uline{M}ultimodal \uline{A}I), that integrates anatomically grounded visual tokens with genetic prompting.
Instead of operating on whole-brain MRI volumes, each T1-weighted scan is parcellated into anatomically defined 3D region-of-interests (ROIs) using a standard atlas, and each ROI is represented as an individual visual token. This preserves regional structural detail and enables the model to attend to localized patterns of neurodegeneration that are relevant to disease staging.
In parallel, SNP sequences are serialized into textual descriptions and encoded as discrete genetic tokens, allowing the LLM to exploit its pre-trained biomedical priors and capture long-range, nonlinear dependencies across variants.
By projecting both ROI tokens and SNP tokens into a shared semantic embedding space, R-GenIMA enables cross-modal self-attention that explicitly models the relationships between genetic variation and spatially specific brain structures.
This design simultaneously bridges the semantic gap between genotype and phenotype and retains the spatial granularity essential for interpreting gene–brain interactions in early Alzheimer’s disease.

% Our methods achieves highly accurate four-class classification across NC, SMC, MCI, and AD on the Alzheimer’s Disease Neuroimaging Initiative (ADNI) dataset \cite{wang2012identifying,mueller2005alzheimer}. 
% To the best of our knowledge, this is the first unified imaging–genetic framework capable of modeling the fine-grained clinical trajectory of early-stage Alzheimer’s disease while simultaneously providing biologically interpretable outputs. 
% Furthermore, the model identifies (i) ROIs with significant stage-specific relevance along the Alzheimer’s disease continuum, (ii) gene segments associated with different disease stages, and (iii) ROI–gene association patterns that reveal multiscale relationships in AD, MCI, and SMC groups. 
We evaluate R-GenIMA on the Alzheimer's Disease Neuroimaging Initiative (ADNI) dataset\cite{wang2012identifying,mueller2005alzheimer}, demonstrating state-of-the-art performance in four-way classification across NC, SMC, MCI, and AD. To the best of our knowledge, this represents the first unified imaging–genetic framework capable of fine-grained disease-stage prediction while simultaneously providing mechanistically interpretable outputs at multiple biological scales. Specifically, the model identifies: (i) stage-specific neuroanatomical signatures that evolve systematically across the AD continuum, (ii) disease-associated genetic markers enriched for established risk loci, and (iii) reproducible ROI–gene associations that reveal coordinated patterns of molecular susceptibility and regional vulnerability. Together, these findings demonstrate that multimodal large language models can bridge micro-scale genetic variation with macro-scale structural brain alterations in a biologically principled and clinically interpretable manner, offering new avenues for early detection, risk stratification, and mechanistic investigation of Alzheimer's disease.
%\textcolor{red}{Our method achieves highly accurate four-class classification across NC, SMC, MCI, and AD within the Alzheimer’s Disease Neuroimaging Initiative (ADNI) dataset \cite{wang2012identifying,mueller2005alzheimer}. To our knowledge, this represents the first unified imaging–genetic framework capable of modeling fine-grained clinical stages of early Alzheimer’s disease while simultaneously yielding biologically interpretable outputs. Beyond prediction, the model also provides rich interpretability by identifying (i) stage-specific ROIs along the Alzheimer’s disease continuum, (ii) genetic segments associated with distinct diagnostic groups, and (iii) ROI–gene association patterns that reveal multiscale relationships across AD, MCI, and SMC.}
\section{Results}\label{sec2}
\begin{table}[ht!]
\caption{Overall classification performance across the three data configurations. We report both Accuracy and Macro-F1 scores. Methods marked with $\dagger$ denote simple feature-concatenation baselines, whereas methods marked with $\ddagger$ refer to cross-modal attention–based fusion models (see Section~\ref{baselines}).}
\centering
\begin{tabular}{|lcc|}
\hline
\textbf{Method} & \textbf{Accuracy} & \textbf{Macro-F1} \\ \hline
\multicolumn{3}{|l|}{\textbf{SNPs-only}} \\ \hline
MLP   & 38.82\% & 26.44\% \\
BERT\cite{devlin2019bert}  & 67.08\% & 43.36\% \\
Llama\cite{touvron2023llamaopenefficientfoundation} & 81.08\% & 65.93\% \\
Qwen\cite{qwen2.5}  & 80.84\% & 66.14\% \\ \hline

% \multicolumn{3}{|l|}{\textbf{Image-only}} \\ \hline
% Med3d & 94.10\% & 87.70\% \\
% 3DViT & 94.84\% & 87.56\% \\ \hline

\multicolumn{3}{|l|}{\textbf{Image-Gene}} \\ \hline
BERT\cite{devlin2019bert}+Med3D\cite{chen2019med3d}$^{\dagger}$      & 94.59\% & 87.16\% \\
BERT\cite{devlin2019bert}+Med3D\cite{chen2019med3d}$^{\ddagger}$  & 90.42\% & 82.45\% \\
BERT\cite{devlin2019bert}+RiT$^{\dagger}$      & 95.09\% & 89.97\% \\
BERT\cite{devlin2019bert}+RiT$^{\ddagger}$  & 92.87\% & 85.67\% \\
Llama\cite{touvron2023llamaopenefficientfoundation}+Med3D\cite{chen2019med3d}              & 93.36\% & 85.71\% \\
Qwen\cite{qwen2.5}+Med3D\cite{chen2019med3d}               & 93.61\% & 85.87\% \\
R-GenIMA (Llama)              & 95.09\% & 90.37\% \\
R-GenIMA (Qwen)               & 95.58\% & 90.42\% \\ \hline

\multicolumn{3}{|l|}{\textbf{Mixture Dataset}} \\ \hline
Llama\cite{touvron2023llamaopenefficientfoundation}+Med3D\cite{chen2019med3d} & 97.54\% & 92.20\% \\
Qwen\cite{qwen2.5}+Med3D\cite{chen2019med3d}  & 96.80\% & 90.23\% \\
R-GenIMA (Llama) & 98.03\% & 97.90\% \\
R-GenIMA (Qwen)  & 99.01\% & 98.50\% \\ \hline
\end{tabular}
\label{tab:overall}
\end{table}

\subsection{Multimodal Classification Performance Across Alzheimer's Disease Stages}
%\subsection{Multimodal Prediction of Fine-Grained Alzheimer’s Disease Stages}
A primary objective of this study is to assess the effectiveness of our proposed multimodal framework for fine-grained Alzheimer’s disease stage classification within the ADNI cohort. 
Specifically, we evaluated: (i) a SNPs-only setting to establish the discriminative capacity of genomic variation; (ii) a paired Image–Gene setting to determine how structural MRI enhances disease-stage representation; and (iii) a Mixture Dataset setting that augments paired samples with additional SNPs-only subjects to examine whether heterogeneous training data can further improve multimodal learning. 
Tables~\ref{tab:overall} summarizes the full set of experiments and corresponding findings.

% Why Multimodal
The results in Table~\ref{tab:overall} first highlight the behavior of genetic predictors in the SNPs-only setting. 
Large language models such as Llama and Qwen achieve accuracies of 81.08\% and 80.84\%, substantially outperforming the two baseline architectures—MLP (38.82\%) and BERT (67.08\%)—demonstrating that LLMs provide a markedly more expressive framework for modeling high-dimensional genomic variation.
Despite this advantage, SNPs-only prediction remains moderate in absolute terms, reflecting a fundamental limitation: genetic variation encodes latent susceptibility rather than expressed pathology, and therefore provides an incomplete view of disease-stage differentiation when used in isolation.
To address this limitation, we introduced structural MRI as a complementary modality and evaluated multimodal imaging–genetics models. Across architectures, integrating imaging features with SNPs embeddings consistently improved performance over the SNPs-only models, confirming that macroscopic neuroanatomical signatures supply critical downstream information that is not present in genomic data alone. For example, the Qwen+Med3D fusion model achieves 93.61\% accuracy, a substantial increase over the 80.84\% achieved by Qwen on genetics alone.

% Why ROI-atlas?
Building on this, we further incorporated our proposed ROI-based visual representation, RiT(ROI-wised Transformer), designed to enhance the multimodal integration mechanism. 
Replacing whole-brain volumes with 3D ROI patches yielded additional performance gains across all fusion architectures. 
For example, the R-GenIMA (Qwen) and R-GenIMA (Llama) models reach 95.58\% and 95.09\% accuracies, consistently outperforming their whole-brain counterparts. 
This improvement suggests that region-level decomposition provides a more discriminative and biologically aligned encoding of structural heterogeneity, enabling the model to more effectively map imaging signals to underlying genetic susceptibility profiles. 
Collectively, these results demonstrate that while genetic predictors alone carry meaningful disease-related information, the combination of macro-scale imaging and micro-scale genomic features—especially under ROI-guided encoding—offers a more powerful and fine-grained representation of early Alzheimer’s disease progression.

% Why mixed data augmentation?
We further examined whether enlarging the genetic sample space could enhance multimodal learning by training on a Mixture Dataset that augments paired imaging–genetic examples with additional SNPs-only subjects. 
This heterogeneous training regime yielded the strongest performance gains observed in our study. The qwen+3droi model reached an accuracy of 99.01\% and a Macro-F1 of 98.50\%, substantially exceeding its performance on the paired Image–Gene dataset (95.58\%). 
This improvement suggests that additional unimodal genetic data refine the structure of the genomic embedding space and strengthen its alignment with imaging-derived features, even though these added subjects lack MRI scans. 
In effect, the model leverages the broader genetic distribution to stabilize disease-stage boundaries and improve cross-modal representation learning, enabling near-perfect discrimination across AD, MCI, SMC, and NC. 
These findings indicate that multimodal imaging–genetics models can benefit meaningfully from heterogeneous datasets and do not require complete modality pairs to achieve state-of-the-art performance.

% Why showing each class prediction?
To further characterize the model’s behavior across disease stages, we additionally report class-specific Precision, Recall, F1-score, and Specificity for NC (see Table~\ref{tab:CN}), SMC (see Table~\ref{tab:SMC}), MCI (see Table~\ref{tab:MCI}) and AD (see Table~\ref{tab:AD}).
Multi-class neurodegeneration classification is intrinsically imbalanced and clinically heterogeneous; therefore, per-class evaluation is essential for determining whether the model maintains consistent performance across heterogeneous subgroups. 
Overall, the model demonstrates stable predictive performance across classes, indicating that it does not rely disproportionately on any single category. Our results reveal clear performance differences across disease stages: predictions for AD and NC achieve higher precision and specificity, reflecting their more distinct imaging signatures, whereas SMC and MCI exhibit comparatively lower recall, consistent with their subtle phenotypes. 

\begin{table}[ht!]
\caption{NC classification performance across different data configurations.
We report Precision, Recall, Macro-F1, and Specificity scores.
Methods marked with $\dagger$ denote simple feature-concatenation baselines, while those marked with $\ddagger$ indicate cross-modal attention–based fusion models (see Section~\ref{baselines}).}
\centering
\begin{tabular}{|lcccc|}
\hline
\textbf{Method} & \textbf{P} & \textbf{R} & \textbf{F1} & \textbf{Specificity} \\ \hline
\multicolumn{5}{|l|}{\textbf{SNPs-only}} \\ \hline
MLP   & 29.90\% & 25.22\% & 27.38\% & 76.71\% \\
BERT\cite{devlin2019bert}  & 53.33\% & 76.52\% & 62.86\% & 73.63\% \\
Llama\cite{touvron2023llamaopenefficientfoundation} & 84.50\% & 86.51\% & 85.49\% & 92.88\% \\
Qwen\cite{qwen2.5}  & 86.89\% & 84.13\% & 85.48\% & 94.31\% \\ \hline

% \multicolumn{5}{|l|}{\textbf{Image-only}} \\ \hline
% Med3d & 91.06\% & 97.39\% & 94.12\% & 96.23\% \\
% 3DViT & 94.57\% & 96.83\% & 95.69\% & 95.71\% \\ \hline

\multicolumn{5}{|l|}{\textbf{Image-Gene}} \\ \hline
BERT\cite{devlin2019bert}+Med3D\cite{chen2019med3d}$^{\dagger}$     & 96.83\% & 96.83\% & 96.83\% & 98.58\% \\
BERT\cite{devlin2019bert}+Med3D\cite{chen2019med3d}$^{\ddagger}$ & 91.41\% & 92.86\% & 92.13\% & 96.09\% \\
BERT\cite{devlin2019bert}+RiT$^{\dagger}$     & 95.35\% & 97.62\% & 96.47\% & 97.86\% \\
BERT\cite{devlin2019bert}+RiT$^{\ddagger}$ & 94.49\% & 95.24\% & 94.86\% & 97.51\% \\
Llama\cite{touvron2023llamaopenefficientfoundation}+Med3D\cite{chen2019med3d}             & 94.74\% & 93.91\% & 94.32\% & 97.94\% \\
Qwen\cite{qwen2.5}+Med3D\cite{chen2019med3d}              & 94.74\% & 93.91\% & 94.32\% & 97.94\% \\
R-GenIMA (Llama)             & 96.75\% & 94.44\% & 95.58\% & 98.85\% \\
R-GenIMA (Qwen)              & 96.85\% & 97.62\% & 97.23\% & 98.58\% \\ \hline

\multicolumn{5}{|l|}{\textbf{Mixture Dataset}} \\ \hline
Llama\cite{touvron2023llamaopenefficientfoundation}+Med3D\cite{chen2019med3d} & 96.64\% & 100\% & 98.29\% & 98.63\% \\
Qwen\cite{qwen2.5}+Med3D\cite{chen2019med3d}  & 99.13\% & 99.13\% & 99.13\% & 99.65\% \\
R-GenIMA (Llama) & 96.58\% & 98.26\% & 97.41\% & 98.63\% \\
R-GenIMA (Qwen)  & 98.28\% & 99.13\% & 98.70\% & 99.31\% \\ \hline
\end{tabular}
\label{tab:CN}
\end{table}
%%%%%%%%%%%%%%%%%%%%%%%%%%%%%%%%%%%%%%%%%%%%%%%%%%%%%%
\begin{table}[ht!]
\caption{SMC classification performance across different data configurations.
We report Precision, Recall, Macro-F1, and Specificity scores.
Methods marked with $\dagger$ denote simple feature-concatenation baselines, while those marked with $\ddagger$ indicate cross-modal attention–based fusion models (see Section~\ref{baselines}).}
\centering
\begin{tabular}{|lcccc|}
\hline
\textbf{Method} & \textbf{P} & \textbf{R} & \textbf{F1} & \textbf{Specificity} \\ \hline
\multicolumn{5}{|l|}{\textbf{SNPs-only}} \\ \hline
MLP   & 7.69\% & 11.76\% & 9.30\% & 93.84\% \\
BERT\cite{devlin2019bert}  & 100\% & 5.88\% & 11.11\% & 100\% \\
Llama\cite{touvron2023llamaopenefficientfoundation} & 13.64\% & 50.00\% & 21.43\% & 95.26\% \\
Qwen\cite{qwen2.5}  & 14.81\% & 66.67\% & 24.24\% & 94.23\% \\ \hline

% \multicolumn{5}{|l|}{\textbf{Image-only}} \\ \hline
% Med3d & 100\% & 52.94\% & 69.23\% & 100\% \\
% 3DViT & 66.67\% & 66.67\% & 66.67\% & 99.50\% \\ \hline

\multicolumn{5}{|l|}{\textbf{Image-Gene}} \\ \hline
BERT\cite{devlin2019bert}+Med3D\cite{chen2019med3d}$^{\dagger}$      & 66.67\% & 66.67\% & 66.67\% & 99.50\% \\
BERT\cite{devlin2019bert}+Med3D\cite{chen2019med3d}$^{\ddagger}$  & 57.14\% & 66.67\% & 61.54\% & 99.25\% \\
BERT\cite{devlin2019bert}+RiT$^{\dagger}$      & 71.43\% & 83.33\% & 76.92\% & 99.50\% \\
BERT\cite{devlin2019bert}+RiT$^{\ddagger}$  & 66.67\% & 66.67\% & 66.67\% & 99.50\% \\
Llama\cite{touvron2023llamaopenefficientfoundation}+Med3D\cite{chen2019med3d}              & 100\% & 47.06\% & 64.00\% & 99.99\% \\
Qwen\cite{qwen2.5}+Med3D\cite{chen2019med3d}               & 100\% & 47.06\% & 64.00\% & 99.99\% \\
R-GenIMA (Llama)              & 71.43\% & 83.33\% & 76.92\% & 99.50\% \\
R-GenIMA (Qwen)               & 71.43\% & 83.33\% & 76.92\% & 99.50\% \\ \hline

\multicolumn{5}{|l|}{\textbf{Mixture Dataset}} \\ \hline
Llama\cite{touvron2023llamaopenefficientfoundation}+Med3D\cite{chen2019med3d} & 100\% & 58.82\% & 74.07\% & 99.99\% \\
Qwen\cite{qwen2.5}+Med3D\cite{chen2019med3d}  & 90.91\% & 58.82\% & 71.43\% & 99.74\% \\
R-GenIMA (Llama) & 100\% & 94.12\% & 96.97\% & 99.99\% \\
R-GenIMA (Qwen)  & 100\% & 94.12\% & 96.97\% & 99.99\% \\ \hline
\end{tabular}
\label{tab:SMC}
\end{table}
%%%%%%%%%%%%%%%%%%%%%%%%%%%%%%%%%%%%%%%%%%%%%%%%%%%%%%
\begin{table}[ht!]
\caption{MCI classification performance across different data configurations.
We report Precision, Recall, Macro-F1, and Specificity scores.
Methods marked with $\dagger$ denote simple feature-concatenation baselines, while those marked with $\ddagger$ indicate cross-modal attention–based fusion models (see Section~\ref{baselines}).}
\centering
\begin{tabular}{|lcccc|}
\hline
\textbf{Method} & \textbf{P} & \textbf{R} & \textbf{F1} & \textbf{Specificity} \\ \hline
\multicolumn{5}{|l|}{\textbf{SNPs-only}} \\ \hline
MLP   & 57.21\% & 53.12\% & 55.09\% & 51.36\% \\
BERT\cite{devlin2019bert}  & 79.37\% & 79.02\% & 79.19\% & 74.86\% \\
Llama\cite{touvron2023llamaopenefficientfoundation} & 89.34\% & 81.11\% & 85.02\% & 88.95\% \\
Qwen\cite{qwen2.5}  & 89.50\% & 82.49\% & 85.85\% & 88.95\% \\ \hline

% \multicolumn{5}{|l|}{\textbf{Image-only}} \\ \hline
% Med3d & 96.41\% & 95.98\% & 96.20\% & 95.63\% \\
% 3DViT & 95.85\% & 95.85\% & 95.85\% & 95.26\% \\ \hline

\multicolumn{5}{|l|}{\textbf{Image-Gene}} \\ \hline
BERT\cite{devlin2019bert}+Med3D\cite{chen2019med3d}$^{\dagger}$     & 93.75\% & 96.77\% & 95.24\% & 92.63\% \\
BERT\cite{devlin2019bert}+Med3D\cite{chen2019med3d}$^{\ddagger}$ & 92.13\% & 91.71\% & 91.92\% & 91.05\% \\
BERT\cite{devlin2019bert}+RiT$^{\dagger}$     & 96.73\% & 95.39\% & 96.06\% & 96.32\% \\
BERT\cite{devlin2019bert}+RiT$^{\ddagger}$ & 94.42\% & 93.55\% & 93.98\% & 93.68\% \\
Llama\cite{touvron2023llamaopenefficientfoundation}+Med3D\cite{chen2019med3d}             & 93.19\% & 97.77\% & 95.42\% & 91.25\% \\
Qwen\cite{qwen2.5}+Med3D\cite{chen2019med3d}              & 93.99\% & 97.77\% & 95.84\% & 92.34\% \\
R-GenIMA (Llama)             & 95.43\% & 96.31\% & 95.87\% & 94.74\% \\
R-GenIMA (Qwen)              & 96.31\% & 96.31\% & 96.31\% & 95.79\% \\ \hline

\multicolumn{5}{|l|}{\textbf{Mixture Dataset}} \\ \hline
Llama\cite{touvron2023llamaopenefficientfoundation}+Med3D\cite{chen2019med3d} & 97.38\% & 99.55\% & 98.45\% & 96.72\% \\
Qwen\cite{qwen2.5}+Med3D\cite{chen2019med3d}  & 96.96\% & 95.55\% & 98.24\% & 96.17\% \\
R-GenIMA (Llama) & 98.65\% & 97.77\% & 98.21\% & 98.36\% \\
R-GenIMA (Qwen)  & 99.55\% & 99.11\% & 99.33\% & 99.45\% \\ \hline
\end{tabular}
\label{tab:MCI}
\end{table}
%%%%%%%%%%%%%%%%%%%%%%%%%%%%%%%%%%%%%%%%%%%%%%%%%%%%%%
%%%%%%%%%%%%%%%%%%%%%%%%%%%%%%%%%%%%%%%%%%%%%%%%%%%%%%
\begin{table}[ht!]
\caption{AD classification performance across different data configurations.
We report Precision, Recall, Macro-F1, and Specificity scores.
Methods marked with $\dagger$ denote simple feature-concatenation baselines, while those marked with $\ddagger$ indicate cross-modal attention–based fusion models (see Section~\ref{baselines}).}
\centering
\begin{tabular}{|lcccc|}
\hline
\textbf{Method} & \textbf{P} & \textbf{R} & \textbf{F1} & \textbf{Specificity} \\ \hline
\multicolumn{5}{|l|}{\textbf{SNPs-only}} \\ \hline
MLP   & 10.53\% & 15.69\% & 12.60\% & 80.89\% \\
BERT\cite{devlin2019bert}  & 38.89\% & 13.73\% & 20.29\% & 96.91\% \\
Llama\cite{touvron2023llamaopenefficientfoundation} & 71.19\% & 72.41\% & 71.79\% & 95.13\% \\
Qwen\cite{qwen2.5}  & 68.97\% & 68.97\% & 68.97\% & 94.84\% \\ \hline

% \multicolumn{5}{|l|}{\textbf{Image-only}} \\ \hline
% Med3d & 90.38\% & 92.16\% & 91.26\% & 96.60\% \\
% 3DViT & 94.55\% & 89.66\% & 92.04\% & 99.14\% \\ \hline

\multicolumn{5}{|l|}{\textbf{Image-Gene}} \\ \hline
BERT\cite{devlin2019bert}+Med3D\cite{chen2019med3d}$^{\dagger}$      & 96.08\% & 84.48\% & 89.91\% & 99.43\% \\
BERT\cite{devlin2019bert}+Med3D\cite{chen2019med3d}$^{\ddagger}$  & 85.71\% & 82.76\% & 84.21\% & 97.71\% \\
BERT\cite{devlin2019bert}+RiT$^{\dagger}$      & 91.23\% & 89.66\% & 90.43\% & 98.57\% \\
BERT\cite{devlin2019bert}+RiT$^{\ddagger}$  & 86.44\% & 87.93\% & 87.18\% & 97.71\% \\
Llama\cite{touvron2023llamaopenefficientfoundation}+Med3D\cite{chen2019med3d}              & 90.00\% & 88.24\% & 89.11\% & 98.59\% \\
Qwen\cite{qwen2.5}+Med3D\cite{chen2019med3d}               & 88.46\% & 90.20\% & 89.32\% & 98.31\% \\
R-GenIMA (Llama)              & 93.10\% & 93.10\% & 93.10\% & 98.85\% \\
R-GenIMA (Qwen)               & 92.86\% & 89.66\% & 91.23\% & 98.85\% \\ \hline

\multicolumn{5}{|l|}{\textbf{Mixture Dataset}} \\ \hline
Llama\cite{touvron2023llamaopenefficientfoundation}+Med3D\cite{chen2019med3d} & 100\% & 96.08\% & 98.00\% & 99.99\% \\
Qwen\cite{qwen2.5}+Med3D\cite{chen2019med3d}  & 92.16\% & 92.16\% & 92.16\% & 98.87\% \\
R-GenIMA (Llama)  & 98.08\% & 100\%  & 99.03\% & 99.71\% \\
R-GenIMA (Qwen)   & 98.08\% & 100\%  & 99.03\% & 99.71\% \\ \hline
\end{tabular}
\label{tab:AD}
\end{table}

\subsection{Model-Prioritized Genes Show Significant Enrichment for Established AD Risk Loci}

We evaluated whether the genes prioritized by the model’s attention rollout–based attribution \cite{abnar2020quantifying} reflect biologically meaningful AD signals rather than arbitrary internal weights. Because our framework operates on a predefined panel of 105 candidate genes, we used this same panel as the background gene universe to ensure a model-consistent enrichment analysis. Within this universe, 45 genes have been previously implicated in AD by genome-wide association studies (GWAS)\footnote{Defined as loci reported with genome-wide significance ($P < 5 \times 10^{-8}$).} based on publicly available summary statistics retrieved from the NHGRI–EBI GWAS Catalog\footnote{Summary statistics downloaded from the NHGRI–EBI GWAS Catalog (\url{https://www.ebi.ac.uk/gwas/}) \cite{10.1093/nar/gkae1070} on 10/02/2025.}, providing a biologically grounded reference set against which to assess enrichment. To derive a robust set of salient genes from the model, we ranked genes by attention-based importance and applied a bootstrap stability procedure involving 1,000 resampling iterations (1,000 samples per iteration), retaining the top 45 genes in each run. Genes appearing in more than 50\% of the iterations were designated as stable high-attention genes, a criterion designed to minimize spurious selections driven by sampling noise or idiosyncratic attention fluctuations. This procedure yielded nine consistently prioritized genes—\textit{RBFOX1}, \textit{TEAM}, \textit{CHAT}, \textit{IGH}, \textit{APOE}, \textit{BIN1}, \textit{CLU}, \textit{CNTNAP2}, and \textit{NECTIN2}—all of which have prior evidence linking them to AD-related pathways.

To quantify whether these model-prioritized genes were disproportionately enriched for established AD risk loci, we constructed a $2 \times 2$ contingency table within the same 105-gene universe and performed a one-sided Fisher’s Exact Test (“greater”), an approach appropriate for small sample sizes and targeted enrichment hypotheses \cite{rivals2007enrichment}. The analysis yielded a statistically significant enrichment (P = 0.027; odds ratio = 5.58). The P-value indicates that, under the null hypothesis that the model’s prioritization is unrelated to GWAS evidence, an overlap as large as the one observed would be expected with probability only $\sim$2.7\%. The odds ratio provides a complementary effect-size interpretation, showing that attention-selected genes are approximately 5.6 times more likely to be GWAS-supported AD genes than non-selected genes within the same feature space. Because the enrichment is tested against the model’s restricted gene universe rather than the whole genome, the analysis is conservative with respect to genome-wide claims; nevertheless, the results demonstrate that the model’s internal attribution aligns strongly with established AD genetics, supporting the biological validity and interpretability of our multimodal imaging–genetics framework.

The enrichment serves as an external validation step, demonstrating that the genes highlighted by the model coincide with independently established GWAS risk loci at a rate significantly above chance. This provides a principled and quantitative indicator of interpretability quality for the attention-based gene ranking within our multimodal AD prediction framework.
% Biologically, the stable high-attention genes include well-known AD-associated loci such as APOE, as well as genes implicated in processes central to AD pathophysiology, including lipid transport and amyloid-related pathways (e.g., APOE, CLU), endocytic and synaptic regulation (e.g., BIN1), neuronal connectivity and synaptic function (e.g., RBFOX1, CNTNAP2, CHAT), and immune-related mechanisms (e.g., IGH). Together, the significant GWAS enrichment and the mechanistic relevance of the prioritized genes suggest that the attention rollout–derived importance scores are not merely artifacts of model optimization, but capture interpretable genetic patterns consistent with known AD risk biology.

% Finally, we note an important boundary: enrichment does not establish causality, and attention is not a direct causal estimator. Rather, this analysis demonstrates external validity of the attribution—i.e., that model-highlighted genes systematically coincide with independently curated GWAS evidence more than expected by chance. This provides a principled, quantitative measure of interpretability quality for the attention-based gene ranking in our multimodal AD prediction framework.

\subsection{Bootstrap-Validated Brain Regions Reveal Stage-Specific Vulnerability Patterns}\label{sec-roi-marker}
\begin{figure*}[ht]
\centering
\includegraphics[width=0.9\textwidth]{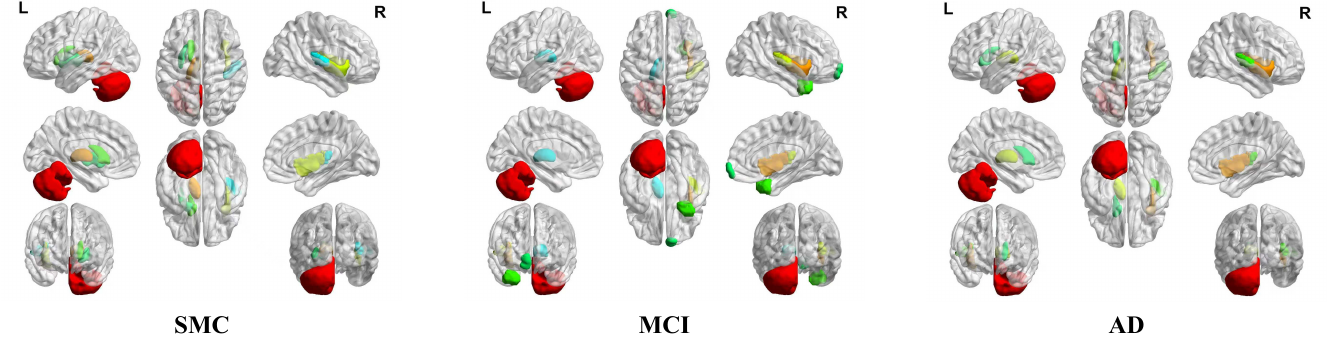}
\caption{Identified brain ROIs associated with disease stages. SMC associated brain ROIs are Left-Cerebellum-Cortex, Left-Thalamus, ctx-rh-insula, Left-Caudate, Left-Putamen and ctx-rh-transversetemporal. MCI associated brain ROIs are Left-Cerebellum-Cortex, ctx-rh-insula, ctx-rh-transversetemporal, ctx-rh-temporalpole, ctx-rh-frontalpole and Left-Thalamus. AD associated brain ROIs are Left-Cerebellum-Cortex, ctx-rh-insula, Left-Thalamus, ctx-rh-transversetemporal and Left-Caudate.}
\label{fig: sig-roi}
\end{figure*}

%\subsection{Key Brain Regions Associated with Disease Stages}\label{sec-roi-marker}
We next applied attention-rollout analysis combined with permutation testing to characterize the neuroanatomical substrates most strongly implicated across disease stages.
Across 1,000 bootstrap iterations, we quantified the reproducibility of regional importance and retained only those ROIs that appeared in more than 50% of runs, ensuring that the reported findings reflect stable and non-spurious signals.
This procedure yielded a compact set of stage-specific loci: six ROIs with consistent relevance in SMC and MCI, and five ROIs prominently associated with AD.
The resulting saliency maps, as shown in Figure \ref{fig: sig-roi}, reveal a structured pattern of shared and divergent neuroanatomical involvement along the Alzheimer’s disease continuum.
Across all three diagnostic groups, the model consistently highlighted a common set of regions—\textit{left cerebellar cortex, right insula, left thalamus,} and \textit{right transverse temporal cortex}—suggesting a core vulnerability architecture detectable even at the earliest symptomatic stages.
Beyond this shared foundation, AD and SMC jointly highlighted the \textit{left caudate}, whereas SMC alone recruited the \textit{left putamen}, indicating selective engagement of striatal circuits during subjective decline.
In contrast, MCI exhibited two additional stage-specific cortical loci—the \textit{right temporal pole} and \textit{right frontal pole}—consistent with emerging frontotemporal network disruption during prodromal impairment \cite{yao2010abnormal,karas2008amnestic,machulda2020cortical}.
Together, these stage-resolved signatures illustrate a progressive reorganization from focal cortical perturbations in SMC, to broader frontostriatal involvement in MCI, and consolidated multimodal network disruption in AD.

% \begin{figure*}[ht]
% \centering
% \includegraphics[width=0.9\textwidth]{images/AD_ROI.jpg}
% \caption{The Key ROI for AD, which includes Left-Cerebellum-Cortex, ctx-rh-insula, Left-Thalamus, ctx-rh-transversetemporal and Left-Caudate}
% \label{fig: AD key roi}
% \end{figure*}

% \begin{figure*}[ht]
% \centering
% \includegraphics[width=0.9\textwidth]{images/mci_roi.jpg}
% \caption{The Key ROI for MCI, which includes Left-Cerebellum-Cortex, ctx-rh-insula, ctx-rh-transversetemporal, ctx-rh-temporalpole, ctx-rh-frontalpole and Left-Thalamus}
% \label{fig: mci key roi}
% \end{figure*}

% \begin{figure}[ht]
% \centering
% \includegraphics[width=0.9\textwidth]{images/smc_roi.jpg}
% \caption{The Key ROI for SMC, which includes Left-Cerebellum-Cortex, Left-Thalamus, ctx-rh-insula, Left-Caudate, Left-Putamen and ctx-rh-transversetemporal}
% \label{fig: smc key roi}
% \end{figure}

%\subsection{Gene--ROI Association Analyses}\label{sec-gene-roi}
\subsection{Reproducible ROI–Gene Associations Across Disease Progression}\label{sec-gene-roi}
\begin{figure}[h]
\centering
\includegraphics[width=1.0\textwidth]{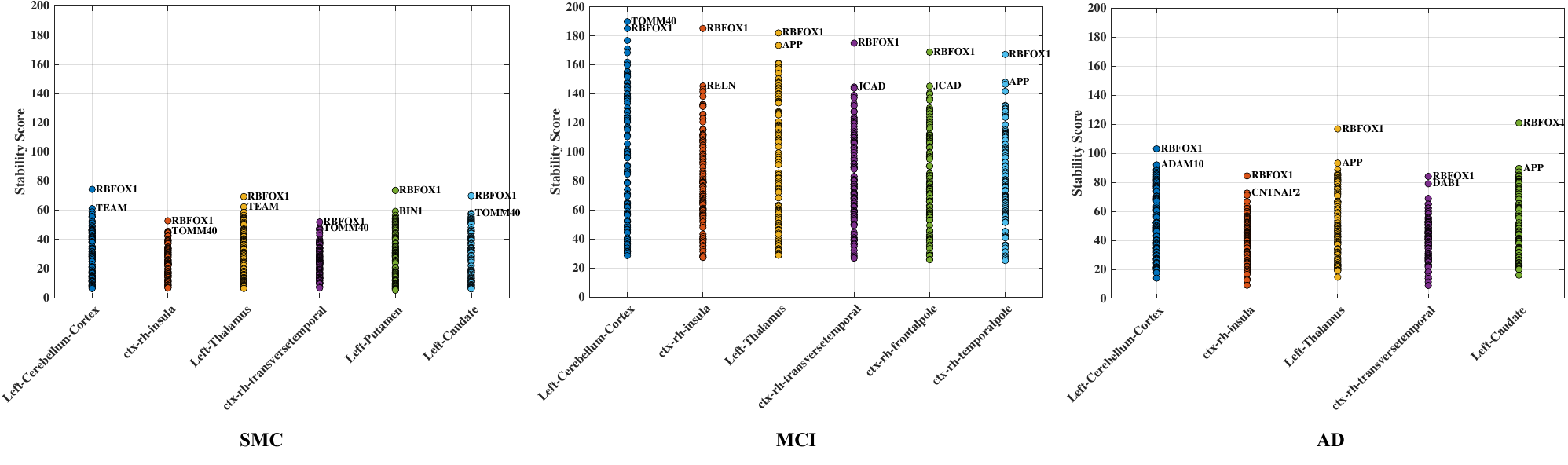}
\caption{Identified ROI–Gene association patterns in the SMC, MCI, and AD cohorts. For each cohort, the two genes with the highest stability scores per ROI are annotated.}
\label{fig: roi-gene-association}
\end{figure}

ROI–gene associations are visualized using a Manhattan-style plot in which ROIs are arranged as contiguous blocks along the x-axis, and each point represents a ROI–gene interaction, colored according to its ROI of origin.
For each ROI, the two most stable gene associations—as determined by the bootstrap-derived stability score—are annotated.
Figure \ref{fig: roi-gene-association} illustrate the resulting association landscapes for the AD, MCI, and SMC cohorts, respectively.
To facilitate clearer interpretation, the visualization is restricted to the disease stage-specific ROIs identified in Section~\ref{sec-roi-marker}.
This representation provides an intuitive summary of the multiscale organization of ROI–gene relationships and highlights distinct, stage-dependent peaks of reproducible genetic involvement.

\section{Discussion}

% \subsection{Architectural Advantages: Overcoming the Dimensionality and Semantic Gaps}
%\subsection{Architectural Advantages: Addressing Multimodal Resolution and Representation Gaps}
\subsection{ROI-Level Decomposition and Semantic Alignment Enable Effective Multimodal Integration}
The superior performance of our multimodal framework, particularly the R-GenIMA (Qwen) variant, reflects its ability to overcome two longstanding challenges in imaging–genetics modeling for early Alzheimer’s disease: the loss of fine-scale neuroanatomical information in conventional 3D imaging pipelines and the representational mismatch between genetic and structural modalities.

First, the ROI-patched strategy offers a principled solution to the resolution–context trade-off inherent to whole-brain 3D CNNs.
Standard volumetric architectures require substantial spatial downsampling to accommodate memory constraints, which can obscure subtle, spatially localized morphological deviations characteristic of prodromal AD.
By decomposing T1 MRI into 109 high-resolution ROI patches ($96^3$ voxels each), our framework preserves regional detail while enabling the Vision Transformer to model distributed inter-regional relationships.
This hierarchical representation retains local structural fidelity yet still captures global brain organization, helping explain the consistent performance improvements over whole-brain approaches in the Image–Gene setting (e.g., 95.58\% vs. 93.61\% for R-GenIMA (Qwen) vs. Qwen+Med3D).

Second, the MLLM architecture mitigates the semantic and statistical mismatch that complicates traditional fusion approaches.
Conventional methods typically employ late or shallow fusion, treating genetic and imaging features as independent numerical descriptors, thereby limiting their ability to capture non-linear genotype–phenotype interactions.
In contrast, our framework embeds SNPs and ROI patches into a unified tokenized representation, enabling cross-modal attention mechanisms to learn how genetic susceptibility patterns modulate regional brain alterations.
The marked performance gap between LLM-based genetic encoders ($\approx$81\% accuracy) and MLP baselines (38.82\%) underscores the value of treating SNPs sequences as structured semantic inputs rather than unordered feature vectors.
This shared representation space allows the model to integrate complementary micro-scale (genetic) and macro-scale (neuroanatomical) information, thereby facilitating biologically meaningful cross-modal reasoning—for example, allowing susceptibility genes such as \textit{APOE}, \textit{BIN1}, or \textit{RBFOX1} to guide attention toward structurally vulnerable ROIs.

\subsection{Heterogeneous Training Data Enhances Multimodal Learning and Cross-Category Robustness}
The empirical findings underscore the value of integrating heterogeneous data sources within the MLLM framework.
The substantial increase in accuracy when training on the Mixture Dataset—where paired imaging–genetic samples are supplemented with additional unimodal genetic data—indicates that the model benefits from broader exposure to genomic variability.
Rather than relying solely on rigid, modality-specific representations typical of CNN-based fusion, the MLLM architecture leverages abundant SNPs-only examples to refine the structure of the genetic embedding space, which in turn stabilizes downstream multimodal learning.
This process effectively transfers population-level susceptibility information from the genetic domain to the imaging domain, enabling the model to interpret structurally ambiguous or subtle neuroanatomical patterns with improved consistency.
Such cross-modal reinforcement appears to be a key contributor to the model’s robustness and its uniformly high performance across diagnostic categories spanning AD, SMC, MCI, and NC.
% The empirical results highlights the synergy of this integration. The leap in accuracy to 99.01\% when training on the Mixed Dataset (augmenting paired data with unimodal examples) suggests that the MLLM architecture possesses superior generalization capabilities compared to rigid CNN-based fusion. The model effectively learns a robust representation of AD pathology by leveraging the vast unimodal genetic data to refine its understanding of susceptibility, which then informs its interpretation of ambiguous imaging features in the multimodal phase. This "knowledge transfer" from abundant genomic data to scarce neuroimaging data is a key driver of the model's state-of-the-art performance across all diagnostic categories (AD, SMC, MCI, NC).

%\subsection{\textcolor{red}{Biological Interpretation of the Identified ROI Markers Across Disease Stages}}
\subsection{Stage-Resolved Neuroanatomical Patterns Align with Established Neurobiology of Early AD}
The stage-specific ROI patterns uncovered by our multimodal framework exhibit strong concordance with contemporary neuroscientific evidence on early Alzheimer’s disease, supporting the biological plausibility of the model’s spatial attributions. 
Notably, four ROIs—left cerebellar cortex, right insula, left thalamus, and right transverse temporal gyrus—are associated with SMC, MCI, and AD, suggesting their role as shared hubs of early vulnerability.
The insula is reported as one of the earliest regions exhibiting functional disconnection and cortical thinning in both SMC and MCI, particularly within salience and interoceptive networks that bridge attention, emotion, and memory processing. Multiple studies demonstrate insular atrophy and network disruption even before objective cognitive impairment manifests \cite{schindler2022associations, hughes2015improving}.
The thalamus, a critical relay hub linking subcortical nuclei with cortico-hippocampal memory systems, exhibits early structural degeneration and impaired connectivity in AD and prodromal stages \cite{zhou2013impaired}.
The transverse temporal gyrus (Heschl’s gyrus) also demonstrates early cortical thinning in AD progression, particularly in SMC and MCI, reflecting its role in multimodal integration and auditory–memory coupling \cite{li2025imaging, wang2020resting}. 
Growing evidence contradicts the traditional view that the cerebellum is largely spared in Alzheimer’s disease and demonstrates cerebellar atrophy and morphological changes associated with cognitive decline in AD and aging \cite{paitel2025cerebellar,tang2021differences,elyan2024hyperactive}

Beyond these shared ROIs, the model also highlights stage-specific distinctions with strong biological interpretability.
Both AD and SMC exhibit involvement of the left caudate, consistent with evidence of early striatal disruptions affecting attention, motivation, and executive function—even before measurable memory deficits arise \cite{xiong2022altered,madsen20103d}
This suggests that striatal vulnerability may precede and potentially contribute to subjective cognitive concerns.
The left putamen, unique to SMC in our findings, has been implicated in very early alterations of motor–cognitive integration and procedural memory, often detectable only in subtle functional changes rather than gross atrophy\cite{wang2020neuroimaging}. 
Its appearance in SMC but not MCI or AD supports the notion that SMC is associated with early basal ganglia involvement before broader cortical deterioration emerges. 
The MCI-associated temporal pole and frontal pole align with well-established cortical biomarkers of prodromal AD. 
Temporal pole thinning marks progression in semantic memory and limbic–default mode network dysfunction \cite{ereira2024early, dickerson2017alzheimer}, while frontal pole deterioration reflects impaired executive control networks often emerging in MCI \cite{friedman2022role,rashidi2020frontal}. 

Taken together, the identified ROIs outline a stage-dependent pattern of neural involvement that is consistent with current understanding of early Alzheimer’s disease while also highlighting a distinct organization of regions emerging from our multimodal analysis.
The ROIs shared across SMC, MCI, and AD correspond to systems involved in salience processing, multimodal integration, and thalamo–cortical communication—domains known to exhibit early physiological vulnerability—whereas the stage-specific ROIs delineate shifts in the dominant neural processes affected across the continuum, from striatal involvement in subjective concerns to association cortical alterations during prodromal impairment.
Notably, the appearance of cerebellar and striatal regions, supported by recent evidence of their involvement in early AD, suggests that the model is sensitive to broader network-level perturbations beyond the classical limbic focus.
The combination and staging of these regions thus constitute a coherent spatial signature emerging from the model, offering additional perspectives on the distributed neural alterations that characterize early Alzheimer’s disease.

% Our attention rollout analysis identified a constellation of pivotal brain regions—including the left cerebellar cortex, right insula, left thalamus, and left caudate—as highly discriminative markers. The identification of these regions is biologically rational and extends beyond the classical focus on the hippocampus.

% \textbf{Subcortical Structures (Thalamus, Caudate)}: The prominence of the caudate and thalamus aligns with emerging evidence that subcortical atrophy and connectivity disruptions occur alongside cortical thinning in AD, often mediating cognitive slowing and executive dysfunction.

% \textbf{Insula}: The right insula's involvement is consistent with the degeneration of the salience network, which is known to be disrupted in early AD and MCI, affecting emotional processing and self-awareness.

% \textbf{Cerebellum}: While traditionally associated with motor control, the detection of the cerebellar cortex supports the "cerebello-cerebral diaschisis" hypothesis, suggesting that cerebellar changes reflect downstream effects of neocortical neurodegeneration.The model's ability to autonomously identify these spatially distributed but functionally connected regions validates its capacity to learn biologically plausible features rather than relying on imaging artifacts.

%\subsection{\textcolor{red}{Biological Interpretation of AD Related Genetic Signals}}
\subsection{Model-Prioritized Genes Span Classical Risk Loci, Synaptic Regulators, and Neuroimmune Pathways}

The nine genes that consistently received high attribution in the AD cohort—\textit{RBFOX1}, \textit{TEAM}, \textit{CHAT}, \textit{IGH}, \textit{APOE}, \textit{BIN1}, \textit{CLU}, \textit{CNTNAP2}, and \textit{NECTIN2}—form a biologically coherent set that spans several molecular pathways implicated in Alzheimer’s disease, while also suggesting interactions that extend beyond the canonical \textit{APOE}-centered framework.
Several of these genes (\textit{APOE}, \textit{BIN1}, \textit{CLU}, \textit{NECTIN2}) have long been recognized as major susceptibility loci through large-scale GWAS, reflecting contributions to lipid metabolism, endocytosis, and amyloid-associated processes \cite{lambert2013meta,kunkle2019genetic}. 
Their prominence in the model aligns with their established relevance to late-onset AD and demonstrates that the multimodal architecture can recover high-confidence risk factors even when operating on a restricted candidate gene set.

Meanwhile, the model highlights a subset of genes—such as \textit{RBFOX1}, \textit{CNTNAP2}, and \textit{CHAT}—that participate in neuronal connectivity, synaptic function, and cholinergic signaling, domains increasingly understood to deteriorate early in AD \cite{chen2022role}.
The joint appearance of synaptic-regulatory genes (\textit{RBFOX1}, \textit{CNTNAP2}) and classical risk loci (\textit{APOE}, \textit{BIN1}) is notable, as it reflects a shift from single-gene interpretation toward multigene network involvement, consistent with the emerging view of AD as a disorder of distributed neural circuitry rather than isolated molecular lesions \cite{kim2022alteration}.
The emergence of immune-related loci such as \textit{IGH} further supports a broader pathophysiological profile, resonating with recent evidence implicating neuroimmune dysregulation in AD progression \cite{leng2021neuroinflammation,hansen2018microglia}.
Importantly, the overall pattern does not merely recapitulate known risk genes but instead reveals a clustered constellation of susceptibility, synaptic, and immune pathways that collectively characterize the AD-stage embeddings learned by the model.
This organization suggests that the attention mechanism could capture coordinated genetic programs that underlie the multiscale imaging signatures observed in the AD cohort, offering a complementary molecular perspective on the disease stage identified through imaging–genetic fusion.

%\subsection{\textcolor{red}{Interpretation of ROI-Gene Associations Across AD Progression}}
\subsection{Brain ROI–Gene Associations Evolve Systematically from Synaptic Vulnerability to Network Disruption}
The stage-dependent organization of Gene–ROI associations uncovered by our model reveals a coherent biological trajectory that aligns with contemporary views of AD as a disorder of progressively disrupted neural circuits. 
Rather than reflecting isolated effects, the associations form structured molecular–anatomical patterns that evolve systematically from SMC through MCI to clinically manifest AD. 
This progression is reflected both in the sets of implicated genes—including \textit{APP}, \textit{RBFOX1}, \textit{TOMM40}, \textit{CNTNAP2}, and \textit{BIN1}—and in the anatomical loci to which they are linked, spanning striatal hubs, thalamic relay nuclei, cerebellar cortices, and multimodal association areas.

In the SMC cohort, stable associations emerged between striatal structures (i.e., caudate, putamen), cerebellar cortex, thalamus, and insula with genetic factors such as \textit{RBFOX1}, \textit{TOMM40}, \textit{TEAM}, and \textit{BIN1}. 
This configuration is biologically reasonable and consistent with recent imaging evidence showing that SMC is accompanied by subtle disruptions in cortico-striatal, salience, and cerebellar networks prior to overt hippocampal degeneration \cite{tsai2025disrupted,cai2020altered}.
\textit{RBFOX1} regulates synaptic splicing and neuronal excitability \cite{gehman2011splicing}, while \textit{BIN1} modulates endocytic trafficking and tau spreading \cite{crotti2019bin1}.
\textit{TOMM40} contributes to mitochondrial integrity, an early-vulnerable process in preclinical AD \cite{wang2020mitochondria,chen2023tomm40,lee2021tomm40}. 
Their appearance in SMC thus suggests that early subjective decline may reflect microcircuit-level vulnerabilities involving synaptic homeostasis, metabolic stress, and endocytic regulation, which manifest preferentially in striatal–cerebellar–salience circuits.

In MCI, the distribution of associations shifts toward association cortex regions—including frontal pole, temporal pole, insula, and transverse temporal gyrus—paired with genes such as \textit{APP}, \textit{RBFOX1}, \textit{TOMM40}, \textit{RELN}, and \textit{JCAD}.
\textit{APP}’s emergence indicates that amyloidogenic processes accelerate during the prodromal stage \cite{hampel2021amyloid,orobets2023amyloid}.
\textit{RBFOX1}’s continued prominence reflects the well-established role of synaptic dysfunction as an early and central mechanism of cognitive impairment.
\textit{TOMM40}’s strong associations further support the hypothesis that mitochondrial dysfunction contributes to early cortical vulnerability in high-energy-demand regions\cite{ferencz2013influence,lee2021tomm40}.
\textit{RELN} and \textit{JCAD} are involved in cortical laminar development and endothelial stability, respectively, both implicated in early AD-related connectivity and microvascular dysfunction \cite{alexander2023reelin,xu2019novel,peter2018jcad}.
This pattern aligns with neuroimaging studies showing that temporal and frontal association cortices undergo measurable atrophy and connectivity loss during MCI, reflecting emerging cognitive deficits \cite{FAN20081731}.
The conjoint appearance of synaptic-regulatory genes and classical AD risk loci indicates that MCI represents a convergence point where molecular susceptibility transitions toward regionally specific cortical degeneration.

In the AD cohort, associations become highly reproducible and anatomically consolidated, involving subcortical hubs (i.e., caudate, thalamus), cerebellar cortex, insula, and transverse temporal cortex. These regions are linked predominantly with \textit{APP}, \textit{RBFOX1}, \textit{CNTNAP2}, \textit{DAB1}, and \textit{NECTIN2}.
\textit{APP} and \textit{NECTIN2} reflect classical amyloid and lipid transport pathways, while \textit{BIN1}/\textit{CLU} family processes continue to underpin endocytic and immune components of pathology \cite{foster2019clusterin,zheng2024Alzheimer}.
\textit{CNTNAP2} contributes to neuronal connectivity and axonal signaling, with mounting evidence indicating early disruption of large-scale communication networks in AD \cite{PENAGARIKANO2011235}.
\textit{DAB1} participates in Reelin signaling, affecting cortical lamination and synaptic stability.
All these processes have been shown to deteriorate notably in AD progression \cite{Iturria-Medina2016}.

These stage-specific Gene–ROI associations outline a coherent molecular–anatomical progression across the Alzheimer’s disease continuum. 
The early involvement of synaptic, metabolic, and endocytic pathways in striatal–cerebellar–salience circuits during SMC, the emergence of amyloidogenic and cortical vulnerability signatures in MCI, and the consolidation of multimodal cortical–subcortical disruptions in AD together suggest that the model captures a biologically structured cascade rather than isolated correlations. 
This organized pattern offers a complementary view of disease progression—one that links genetic susceptibility to regionally specific network vulnerability—and underscores the potential of multimodal learning frameworks to reveal interpretable, mechanistically informed signatures of neurodegeneration.

\subsection{Translational Potential and Current Limitations of the Multimodal Framework}
Our proposed R-GenIMA framework has several potential clinical impacts.
By integrating structural MRI with genetic variation in an interpretable architecture, the model identifies stage-dependent molecular–anatomical signatures that may enable earlier and more biologically informed detection of individuals at risk for cognitive decline. 
Such signatures could support precision stratification in clinical trials, guide hypotheses about pathway-specific interventions, and help link genetic susceptibility to regionally specific patterns of neurodegeneration. 
The ability to generate consistent, mechanistically plausible Gene–ROI associations across SMC, MCI, and AD further suggests that multimodal learning may provide clinically meaningful insight into how heterogeneous biological processes shape disease trajectories.

Despite these promising findings, several important limitations warrant consideration. Most fundamentally, our analyses are constrained to the ADNI cohort, which, while well-characterized and widely used in AD research, represents a research-recruited population that may not fully capture the demographic, genetic, and clinical heterogeneity encountered in real-world healthcare settings. ADNI participants are predominantly non-Hispanic White individuals with relatively high educational attainment, potentially limiting generalizability across diverse populations with different genetic architectures, environmental exposures, and patterns of brain aging. External validation in independent cohorts—particularly those with broader representation across ancestry groups, socioeconomic backgrounds, and geographic regions—will be essential to assess whether the model's learned associations generalize beyond this specific dataset. Additionally, because ADNI remains one of the few large-scale studies providing paired structural MRI and comprehensive genetic data, opportunities for such validation are currently limited, underscoring the need for continued investment in multimodal data collection efforts.

From a methodological perspective, our framework operates on a predefined candidate gene panel (105 genes, yielding 1,079 SNPs after quality control) selected based on prior AD literature. While this targeted approach enhances interpretability and computational efficiency, it inherently restricts the model's ability to discover novel genetic associations outside this curated set. Future work incorporating genome-wide SNP data or whole-genome sequencing could reveal additional susceptibility loci not captured by our current gene panel, though such expansion would require careful consideration of increased dimensionality and multiple testing concerns. Similarly, our ROI-based parcellation strategy, while preserving regional anatomical detail, relies on a fixed atlas (Desikan-Killiany) that may not optimally capture disease-relevant boundaries, particularly in subcortical structures or regions undergoing subtle morphological changes during preclinical stages.

Finally, while our study demonstrates that interpretable multimodal AI can synthesize imaging and genetic data to produce biologically coherent disease signatures, translation from research findings to clinical implementation requires addressing additional challenges beyond model performance. These include establishing standardized preprocessing pipelines compatible with clinical imaging protocols, developing user interfaces that convey probabilistic predictions and uncertainty estimates appropriately, ensuring robustness to variation in scanner hardware and acquisition parameters, and conducting prospective validation studies to evaluate real-world utility and potential impact on patient outcomes. The current work provides a methodological foundation and proof-of-concept demonstration, but the pathway from research tool to clinically actionable system will require sustained collaborative effort among computational scientists, neuroimaging researchers, geneticists, and clinicians.

Overall, these findings illustrate both the promise and current boundaries of multimodal interpretable learning in Alzheimer's research. The R-GenIMA framework successfully bridges micro-scale genetic variation with macro-scale structural brain alterations in a biologically principled manner, offering new opportunities for early detection, mechanistic investigation, and precision medicine approaches to AD. As multimodal datasets expand and validation efforts progress, such approaches may ultimately contribute to more personalized, biology-informed strategies for identifying at-risk individuals and guiding therapeutic development.
%\textcolor{red}{Because ADNI remains one of the few large-scale datasets providing paired structural MRI and genetic data, our analyses were necessarily constrained to this cohort. Although ADNI is a well-characterized research dataset, broader validation in future multimodal cohorts—once such resources become available—will be essential for fully assessing population-level generalizability.} Overall, these findings illustrate both the promise and the current boundaries of multimodal, interpretable learning in Alzheimer’s research, offering a foundation for future studies aimed at translating imaging–genetic signatures into clinically actionable tools.
\section{Methods}

\subsection{Preliminary Konwledge}

\textbf{Self-Attention Mechanism.}
Self-attention is the core component of Transformer architectures \cite{vaswani2017attention}, designed to model long-range dependencies by relating different positions within a sequence. Unlike recurrent neural networks (RNNs), which process tokens sequentially, self-attention operates on all tokens in parallel.

Given an input sequence $\mathbf{X} \in \mathbb{R}^{n \times d}$, three learnable linear projections are applied to obtain the Query ($\mathbf{Q}$), Key ($\mathbf{K}$), and Value ($\mathbf{V}$) matrices:
\begin{equation}
\mathbf{Q} = \mathbf{X}\mathbf{W}^Q, \quad
\mathbf{K} = \mathbf{X}\mathbf{W}^K, \quad
\mathbf{V} = \mathbf{X}\mathbf{W}^V .
\end{equation}

Attention weights are computed using scaled dot-product attention:
\begin{equation}
\text{Attention}(\mathbf{Q}, \mathbf{K}, \mathbf{V}) =
\text{softmax}\left(\frac{\mathbf{Q}\mathbf{K}^\top}{\sqrt{d_k}}\right)\mathbf{V},
\end{equation}
where $d_k$ denotes the key dimension, and the scaling factor stabilizes training by preventing large dot-product values from saturating the softmax.

\textbf{Transformers.}
Building upon self-attention, Transformers \cite{vaswani2017attention} consist of stacked layers combining multi-head self-attention (MSA) and feed-forward networks (FFN), interconnected via residual connections and layer normalization. Depending on the task, Transformers can adopt encoder-only, decoder-only, or encoder--decoder architectures.

BERT \cite{devlin2019bert} is an encoder-only Transformer that learns bidirectional representations through a masked language modeling objective, where randomly masked tokens are predicted from surrounding context. In bioinformatics, BERT-style models have been adapted to genetic sequences \cite{le2021transformer} by treating DNA segments as sentences and variants (e.g., SNPs) as tokens, enabling the modeling of long-range epistatic interactions across genetic loci.

The Vision Transformer (ViT) \cite{dosovitskiy2021an} extends Transformers to images by partitioning an image into fixed-size patches that are linearly embedded as tokens. Learnable positional embeddings preserve spatial structure, allowing global context modeling across the image. This property is particularly beneficial for medical imaging, where long-range spatial dependencies (e.g., inter-hemispheric brain symmetry) are clinically relevant and difficult for CNNs with local receptive fields to capture.

\textbf{Cross-Modal Attention Mechanism.}
While self-attention models relationships within a single modality, cross-modal attention (CMA) \cite{wei2020multi} enables information fusion across different modalities (e.g., imaging and genetics). In CMA, one modality provides Queries ($\mathbf{Q}$), while the other supplies Keys ($\mathbf{K}$) and Values ($\mathbf{V}$).

For example, aligning genetic features $\mathbf{F}_{gene}$ with image features $\mathbf{F}_{img}$ can be formulated as:
\begin{equation}
\text{CrossAttention}(\mathbf{F}_{gene}, \mathbf{F}_{img}) =
\text{softmax}\left(
\frac{(\mathbf{F}_{gene}\mathbf{W}^Q)(\mathbf{F}_{img}\mathbf{W}^K)^\top}{\sqrt{d_k}}
\right)(\mathbf{F}_{img}\mathbf{W}^V).
\end{equation}

This operation computes a weighted aggregation of image features conditioned on genetic information, allowing genetic risk factors to selectively attend to brain regions most relevant to disease pathology.

\textbf{Convolutional Neural Networks.}
Convolutional Neural Networks (CNNs) are specialized architectures for grid-structured data such as images. They extract hierarchical representations through convolutional filters that capture local patterns (e.g., edges and textures), followed by pooling operations for spatial downsampling.

Med3D \cite{chen2019med3d} is a 3D CNN pre-trained on large-scale, heterogeneous medical volumes to address data scarcity in medical imaging. By learning domain-invariant volumetric representations across modalities (e.g., MRI and CT), Med3D serves as a strong backbone for encoding 3D brain MRI, outperforming models trained from scratch or transferred from natural images.

\subsection{R-GenIMA Architecture}
\begin{figure}[h]
\centering
\includegraphics[width=1.0\textwidth]{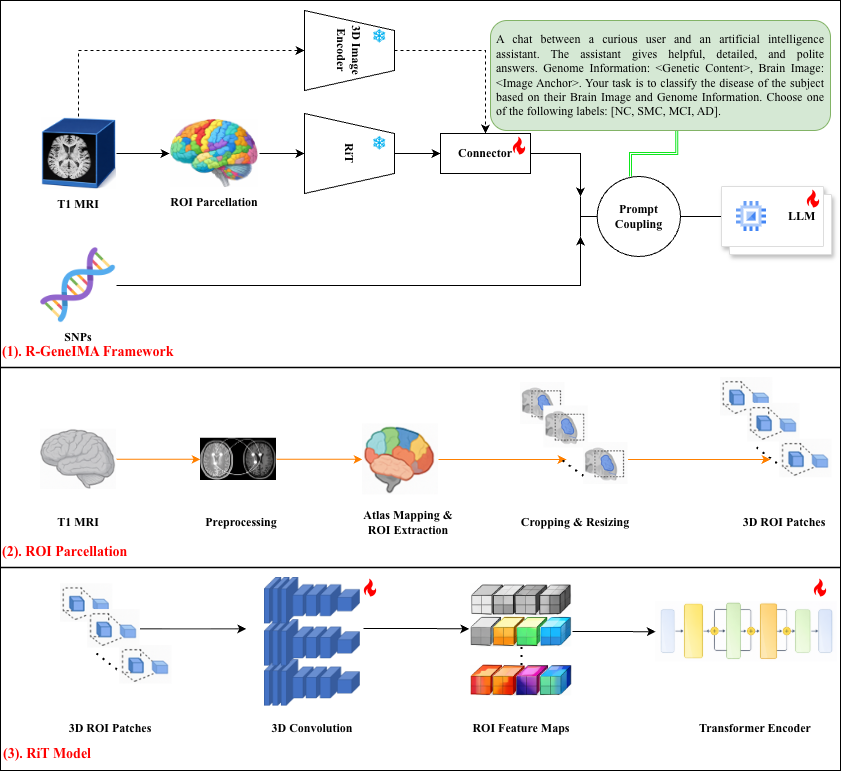}
\caption{(1) The R-GeneIMA framework integrates ROI-based neuroimaging features with SNPs profiles. ROI patches are embedded using the proposed RiT model, and a structured prompt couples these embeddings with the SNPs profile before being passed to the LLM for stage-specific reasoning.
(2) Illustration of the 3D ROI parcellation process.
(3) Illustration of the RiT model.}
\label{fig:main}
\end{figure}
Our proposed ROI–Gene fused Interpretable Multimodal AI framework (R-GenIMA) is illustrated in Figure~\ref{fig:main}.
Given a subject with a T1-weighted 3D brain MRI volume denoted as $\mathbf{I} \in \mathbb{R}^{H \times W \times D \times C}$, and a corresponding genetic SNPs profile denoted by $\mathbf{G}$, R-GenIMA encodes the 3D parcellated anatomical regions using our designed ROI-wise vision Transformer (RiT) model and embeds the SNPs profile using a structured prompting strategy. 
To enable controlled comparisons, the RiT model can be substituted with a whole-brain 3D MRI encoder to process the unsegmented MRI volume, serving as a baseline for evaluating the contribution of ROI-level decomposition. 

% In this section, we introduce our proposed RiT and R-GenIMA frameworks. As illustrated in Figure \ref{fig: roi-gene framework}, the pipeline consists of three stages: (1) ROI Extraction, where 3D Regions of Interest (ROIs) are extracted from the original T1-weighted images; (2) 3D Encoder Pretraining, where we present two distinct approaches for comparison—a whole-brain baseline and our proposed RiT method; and (3) SFT Fine-tuning, where visual features are integrated with genetic SNP data to fine-tune the Multimodal Large Language Model (MLLM). Given that we utilize two distinct vision encoders, we define the specific combination of RiT and the LLM as R-GenIMA (ROI Genome fused Interpretable Multimodal AI).

% Given a subject with a 3D T1-weighted brain MRI volume $\mathbf{X} \in \mathbb{R}^{H \times W \times D \times C}$ and a set of genetic SNPs \textcolor{red}{string including SNP information} denoted by $\mathbf{G}$, our framework employs a dual-pathway strategy to encode visual features and a prompt-based strategy to encode genetic data.

\subsubsection{RiT Model and Image Embedding}
To preserve fine-grained anatomical information while avoiding the memory constraints inherent in whole-brain volumetric processing, we introduce an ROI-wise Vision Transformer (RiT) that operates on anatomically parcellated 3D patches.  
Each T1-weighted MRI volume $\mathbf{I}$ is first segmented into atlas-defined ROIs:
\begin{equation}
    \{\mathbf{I}_{r}^{(i)}\}_{i=1}^{N} = \text{Segment}(\mathbf{I}),
\end{equation}
where $\mathbf{I}_{r}^{(i)}$ denotes the $i$th ROI volume.  
For consistency across subjects, we fix the number of ROIs to $N=109$ (the maximal count in the dataset) and resample each ROI to a standardized spatial dimension of $96 \times 96 \times 96$.  
Missing or anatomically absent ROIs are filled with zero-valued patches, ensuring a uniform input sequence.

Each $96^3$ ROI patch is embedded using a 3D convolutional patch encoder whose kernel spans the entire spatial extent of the ROI and is applied with a stride of 1 along the ROI sequence dimension, producing a single visual token per region:
\begin{equation}
    \mathbf{z}_{r}^{(i)} = \text{Conv3D}(\mathbf{I}_{r}^{(i)}) \in \mathbb{R}^{1 \times 768}.
\end{equation}
These token embeddings are subsequently refined by a lightweight Vision Transformer, which models inter-ROI dependencies and extracts region-specific structural signatures:
\begin{equation}
    \mathbf{h}_{r}^{(i)} = \text{ViT}(\mathbf{z}_{r}^{(i)}).
\end{equation}
Finally, the resulting sequence of ROI embeddings is aligned to the large language model (LLM) embedding space through a linear connector:
\begin{equation}
    \mathbf{H}_{\text{image}} = \text{Linear}\big(\{\mathbf{h}_{r}^{(i)}\}_{i=1}^{N}\big).
\end{equation}
This produces a set of visual tokens compatible with the LLM’s multimodal attention mechanism, enabling direct cross-modal interactions with SNPs-derived genetic tokens.

As a comparative reference, we also include a whole-brain MRI encoder that captures global patterns of structural atrophy. 
In this branch, the full 3D T1 volume $\mathbf{I}$ is processed using a pre-trained Med3D architecture, a 3D convolutional neural network optimized for medical volumetric feature extraction. 
The model produces a single global representation summarizing large-scale anatomical structure:
\begin{equation}
    \mathbf{h}_{wb} = \text{Med3D}(\mathbf{I}) \in \mathbb{R}^{1 \times 256}.
\end{equation}
This global feature vector is subsequently projected into the LLM embedding space through a linear transformation:
\begin{equation}
    \mathbf{H}_{image} = \text{Linear}(\mathbf{h}_{wb}).
\end{equation}
This baseline provides a contrast to the region-wise RiT encoder by emphasizing coarse whole-brain morphology rather than localized, high-resolution anatomical patterns.

\subsubsection{Prompt Design for Imaging–SNPs Coupling}\label{sec:prompt-design}
To leverage the cross-modal reasoning capabilities of the MLLM, we design a structured instruction prompt that jointly incorporates genetic information and MRI-derived visual embeddings within a unified semantic context. 
The prompt enables the model to align textual representations of genetic variation with spatially localized neuroanatomical features, thereby supporting token-wise cross-attention across modalities.  
The template is defined as follows:
\fbox{%
  \parbox{1.0\linewidth}{%
    \small\ttfamily{A chat between a curious user and an artificial intelligence assistant. 
    The assistant gives helpful, detailed, and polite answers. 
    Genome Information: $<$Genetic Content$>$ 
    Brain Image: $<$Image Content$>$. 
    Your task is to classify the disease of the subject based on their Brain Image and Genome Information. 
    Choose one of the following labels: [NC, SMC, MCI, AD].}
  }%
}
\noindent Multimodal coupling is enabled through a two-level insertion mechanism:\\
\noindent\uline{Text-level genetic insertion.}
The placeholder {\ttfamily{$<$Genetic Content$>$}} is replaced with a serialized textual representation of the subject’s SNPs profile. 
This converts discrete allelic values into natural-language sequences that the LLM can interpret using its biomedical pretraining, enabling contextual reasoning over genetic patterns.

\noindent\uline{Embedding-level visual insertion.}
The placeholder {\ttfamily{$<$Image Content$>$}} acts as a dedicated anchor token. 
During the forward pass, it is replaced with the sequence of visual embeddings produced by the ROI-wise 3D vision encoder. 
These embeddings are mapped into the LLM token space by the Connector module, allowing the model to perform cross-attention between specific genetic variants and anatomically localized brain features.
The MLLM receives this combined prompt as input and generates a natural-language diagnostic output of the form:
\fbox{%
  \parbox{1.0\linewidth}{%
    \small\ttfamily{This subject is $<$label$>$.}
  }%
}

\subsubsection{Generative Reasoning via the MLLM}
For the genetic modality, the SNPs sequence $\mathbf{G}$ is serialized into natural-language text and incorporated into the instruction prompt as explained in Section \ref{sec:prompt-design}. 
The tokenizer maps this text into an input token sequence 
$X = \{x_1, x_2, \dots, x_n\}$ containing both the instructional template and the genetic content, together with an image-anchor placeholder that will later be substituted by visual embeddings. 
The diagnostic output is expressed as a target token sequence 
$Y = \{y_1, y_2, \dots, y_m\}$.
The MLLM models the conditional probability of $Y$ given both the textual input and the image-derived embeddings $\mathbf{H}_{image}$ via an autoregressively  factorized joint probability:
\begin{equation}
    P_\theta(Y \mid X, \mathbf{H}_{image})
    = \prod_{t=1}^{m} P_\theta\bigl(y_t \mid y_{<t}, X, \mathbf{H}_{image}\bigr),
\end{equation}
where $y_{<t}$ denotes previously generated tokens. 
The term $\mathbf{H}_{image}$ represents the fused ROI-level visual tokens projected into the LLM embedding space, enabling multimodal cross-attention during generation.

\noindent\textbf{Optimization.}
Model parameters $\theta$ are optimized by minimizing the negative log-likelihood (NLL) over a batch of size $B$:
\begin{equation}
    \mathcal{L}(\theta)
    = -\frac{1}{B} \sum_{j=1}^{B} \sum_{t=1}^{m_j}
      \log P_\theta\!\left(
        y_t^{(j)} \mid 
        y_{<t}^{(j)}, 
        X^{(j)}, 
        \mathbf{H}_{image}^{(j)}
      \right).
\end{equation}

\noindent\textbf{Inference.}
During inference, the fine-tuned model $\mathrm{LLM}_{\theta^*}$ autoregressively predicts the diagnosis token by token:
\begin{equation}
    \hat{y}_t = \arg\max_{v \in \mathcal{V}}
    P_{\theta^*}\!\left(
        v \mid \hat{y}_{<t}, X, \mathbf{H}_{image}
    \right),
\end{equation}
where $\mathcal{V}$ denotes the vocabulary.
This generative formulation allows the model to integrate textual SNPs patterns with anatomically localized MRI embeddings through cross-modal attention, enabling semantically aligned reasoning over multimodal biological inputs.

\subsection{Cluster-based Attention Stability Analysis for ROI--Gene Associations}
To identify robust and biologically meaningful interactions between genetic factors and regional neuroanatomy across disease stages, we developed a bootstrap-based attention stability framework. 
For each subject $s$, the multimodal model produces an attention weight $a_{s,r,g}$ that reflects the contribution of gene $g$ to the representation of region of interest (ROI) $r$. 
For every ROI--gene pair $(r,g)$, these values are aggregated across subjects within a diagnostic group (SMC, MCI, or AD) to form
\begin{equation}
    X_{r,g} = \{a_{1,r,g}, a_{2,r,g}, \ldots, a_{N,r,g}\},
\end{equation}
which represents the empirical distribution of interaction strengths for that group.
Because raw attention estimates may be sensitive to inter-subject variability and occasional outliers, we quantified the reproducibility of each ROI--gene association via nonparametric bootstrap resampling. 
For each pair $(r,g)$, we generated $B = 1000$ bootstrap replicates by sampling $N$ values from $X_{r,g}$ with replacement, and computed a mean attention value for each replicate:
\begin{equation}
    \mu_{r,g}^{(b)} = \frac{1}{N} \sum_{i=1}^{N} x_{i}^{(b)}.
\end{equation}
This process yields an empirical distribution
\begin{equation}
    \{\mu_{r,g}^{(1)}, \mu_{r,g}^{(2)}, \ldots, \mu_{r,g}^{(B)}\},
\end{equation}
capturing both the central tendency and variability of the ROI--gene interaction.
From this distribution, we compute the bootstrap mean
\begin{equation}
    \bar{\mu}_{r,g} = \frac{1}{B} \sum_{b=1}^{B} \mu_{r,g}^{(b)}
\end{equation}
and derive a 95\% confidence interval (CI) from the 2.5th and 97.5th percentiles. 
To summarize association robustness, we define a stability metric:
\begin{equation}
    \text{Stability}_{r,g} 
    = \frac{\bar{\mu}_{r,g}}
           {\text{CI}_{97.5\%} - \text{CI}_{2.5\%}},
\end{equation}
which normalizes the average attention strength by its uncertainty. 
Higher stability values correspond to associations that are both strong and consistently recovered across subjects, allowing prioritization of ROI--gene pairs most likely to reflect disease-relevant biological pathways.

\subsection{Dataset}
\subsubsection{Data Description and Preprocessing}
\begin{table}[t]
\centering
\caption{Simplified summary of ADNI diagnostic criteria for NC, SMC, MCI, and AD.}
\label{tab:diagnosis_criteria}
\begin{tabular}{p{1.1cm} p{3.4cm} p{2.7cm} p{3.6cm}}
\hline
\textbf{Category} & \textbf{Cognitive Tests} & \textbf{Functional Status} & \textbf{Key Features} \\
\hline
\textbf{NC} &
MMSE 24--30; CDR = 0; LM-II normal &
No functional impairment &
No subjective complaints; cognitively normal. \\[4pt]
\textbf{SMC} &
MMSE 24--30; CDR = 0; normal scores &
No impairment &
Subjective memory concerns without objective deficits. \\[4pt]
\textbf{MCI} &
MMSE 24--30; CDR = 0.5; LM-II below cutoff &
Minimal impairment; ADLs preserved &
Objective memory impairment but not dementia. \\[4pt]
\textbf{AD} &
MMSE 20--26; CDR 0.5--1 &
Functional decline &
Meets NINCDS–ADRDA criteria for probable AD. \\
\hline
\end{tabular}
\end{table}

We utilize 3D T1 MRI and corresponding SNPs data from the publicly available Alzheimer’s Disease Neuroimaging Initiative (ADNI) study \cite{alzheimer2010adni}. 
The dataset includes 1998 subjects, consisting of 644 NC (mean age = 73.29 $\pm$ 6.16, 349 female), 100 SMC (mean age = 72.33 $\pm$ 5.67, 59 female), 916 MCI (mean age = 73.06 $\pm$ 7.57, 370 female), and 338 AD (mean age = 74.99 $\pm$ 7.92, 149 female). 
Brief diagnostic criteria for NC, SMC, MCI, and AD used in this study are summarized in Table~\ref{tab:diagnosis_criteria}. 
Full clinical definitions and assessment protocols are provided in the ADNI Procedures Manual and related diagnostic guidelines \cite{weiner2013alzheimer,aisen2010clinical}.

Processing of 3D T1-weighted MRI data was performed using the standard FreeSurfer pipeline (recon-all, version 6.0)\cite{FISCHL2012774}. For each subject, recon-all was applied to carry out skull stripping, intensity normalization, cortical surface reconstruction, and automated segmentation of cortical and subcortical structures. Regional definitions were obtained from the Desikan–Killiany cortical parcellation\cite{desikan2006automated} and FreeSurfer’s automated subcortical segmentation. These atlas-based labels were used to define anatomically consistent ROIs across subjects for subsequent analysis.

The SNPs quality-control procedure consisted of three standard filtering steps: handling missingness, filtering by minor allele frequency (MAF), and testing for deviation from Hardy–Weinberg equilibrium (HWE).
First, SNPs with more than 95\% missing genotypes were removed; for the remaining loci, missing values were imputed using the empirical allele-frequency–based expected genotype.
Second, variants with a MAF below 0.05 were excluded to avoid instability in downstream modeling.
Third, SNPs exhibiting significant deviation from HWE (p-value $< 1 \times 10^{-6}$) were removed, as such deviations may reflect technical artifacts or population substructure.
After applying these quality-control criteria, 1,079 SNPs remained from the original 76K variants.

\subsubsection{Genetic Data Augmentation via Gene-Order Permutation}
Since individual genes do not possess an inherently meaningful sequential ordering in the context of variant–disease prediction, the model should, in principle, exhibit permutation invariance to the arrangement of gene blocks. To enforce this inductive bias and simultaneously expand the effective training distribution, we implemented a gene-order permutation augmentation strategy.

For each subject, the SNPs sequences belonging to a given gene (e.g., $SNP_{i_1}, \dots, SNP_{i_K}$ for $Gene_i$) were preserved as an intact block, while the ordering of these gene-level blocks (e.g., $Gene_i$, $Gene_j$, $\dots$) was randomly permuted. Critically, the internal SNPs order within each gene was not altered, ensuring that biologically meaningful within-gene structure was retained. This augmentation yields alternate yet semantically equivalent genetic representations, promoting robustness to superficial input permutations.

Through repeated random permutations, the dataset was substantially expanded to 50{,}000 training samples and 10{,}000 testing samples. This strategy serves two complementary purposes: (i) reinforcing the model’s invariance to arbitrary gene ordering, thereby preventing spurious sequence-dependent learning, and (ii) increasing genetic sample diversity to improve generalization in downstream multimodal classification.

\subsubsection{Experimental Data Configurations}
Following genetic augmentation and ROI-based image processing, we constructed three complementary data configurations to systematically evaluate the contributions of unimodal and multimodal information, as well as the impact of incorporating heterogeneous samples during model training:
\begin{enumerate}
\item \textbf{Gene-only Configuration}:
This configuration contains exclusively SNPs sequences (after gene-order permutation augmentation).
It isolates the predictive value of genetic variation and serves as a baseline for evaluating the MLLM’s capacity to model high-dimensional genomic signals without any imaging input.
\item \textbf{Image--Gene Configuration}:  
Each sample includes both the subject’s SNPs sequence and their corresponding 3D T1-weighted MRI scan.  
This paired multimodal dataset enables evaluation of cross-modal reasoning and quantifies the benefits of joint imaging–genetic representation learning. Only subjects with complete genetic and imaging data are included.
\item \textbf{Mixture Data Configuration}:  
To examine whether unimodal genetic samples can enhance multimodal learning, we constructed a hybrid dataset in which 50\% of samples contain paired SNPs–MRI data, while the remaining 50\% contain SNPs sequences only.  
This configuration increases genetic sample diversity and allows the model to refine genomic embeddings even when imaging data are unavailable, thereby testing the hypothesis that additional genetic-only samples can improve downstream multimodal classification.
\end{enumerate}
Together, these three data configurations enable a structured evaluation of (i) the stand-alone utility of genetic predictors, (ii) the advantages of multimodal fusion, and (iii) the effect of integrating heterogeneous genetic-only samples into the training of a unified multimodal MLLM framework.

\subsection{Implementation Details}
\subsubsection{Baseline Models}
\label{baselines}

To comprehensively evaluate the performance of our proposed framework, we compared it against a set of representative unimodal and multimodal baselines spanning classical neural architectures, pretrained language models, and standard fusion strategies.

\textbf{Gene-only Baselines.}
These models isolate the predictive contribution of genetic information. Three families of text-based encoders were considered:

\textit{MLP} \cite{rumelhart1986learning}:  
A conventional feedforward neural network that treats SNPs vectors as fixed-length numerical inputs, serving as a non-sequential learning baseline.

\textit{BERT} \cite{devlin2019bert}:  
A bidirectional transformer encoder applied to serialized SNPs sequences to capture local contextual dependencies, representing a discriminative language-modeling approach.

\textit{Large Language Models (Llama \cite{touvron2023llamaopenefficientfoundation} and Qwen \cite{qwen2.5})}:  
Two state-of-the-art generative LLMs—Llama3-8B-Instruct and Qwen2.5-7B—were employed to assess the benefit of large-scale pretraining for interpreting genetic descriptions in the absence of imaging information.

\textbf{Multimodal (Image–Gene) Baselines.}
To benchmark our ROI-guided fusion approach against traditional multimodal pipelines, we paired a 3D medical imaging encoder with a genetic text encoder.

\textit{BERT + Med3D \cite{chen2019med3d} (concat)}:  
A late-fusion architecture in which global whole-brain features extracted by Med3D (a 3D ResNet-based backbone) are concatenated with BERT-derived genetic embeddings prior to classification.

\textit{BERT + Med3D (crossmodal)}:  
A cross-attention baseline that models interactions between whole-brain image embeddings and textual SNPs representations using standard multimodal attention layers.

Together, these baselines allow us to disentangle the individual contributions of genetic feature modeling (MLP vs. BERT vs. LLMs) and to directly compare conventional fusion strategies against the proposed ROI-guided, semantically aligned multimodal integration.

% \subsubsection{Experimental Setting.}
% % Two pretrained models are selected to conduct our experiments, including Llama3-8B-instruct \cite{touvron2023llamaopenefficientfoundation}, Qwen2.5-7B \cite{qwen2.5}. 
% We first train the vision encoders, including Med3D and RiT, on brain MRI classification, where each image is classified into one of four categories: NC, SMC, MCI, or AD. After this stage, the vision encoders are frozen, and the connector and large language model (LLM) are jointly fine-tuned. The fine-tuning is performed on 
% 8× A100 GPUs using full-parameter training with the AdamW optimizer. We employ DeepSpeed ZeRO-3 to parallelize the model parameters, training batches, and optimizer states. The batch size on each GPU is 16. All parameters are trained in bfloat16 precision with a learning rate of $2*10^{-6}$. The models are trained for 3 epochs, and fine-tuning each model requires approximately 48 hours.
\subsubsection{Experimental Setting.}
We adopt a two-stage training strategy. First, the vision encoders—Med3D and the proposed RiT—are trained on a four-way brain MRI classification task (NC, SMC, MCI, AD). After convergence, the encoder weights are frozen. In the second stage, the Connector and the LLM are jointly fine-tuned while receiving fixed visual embeddings from the pretrained encoders.

Fine-tuning is conducted on 8$\times$A100 GPUs using full-parameter training with the AdamW optimizer. We employ DeepSpeed ZeRO-3 to shard model parameters, gradients, and optimizer states across devices. Each GPU processes a batch of 16 samples. All parameters are trained in \texttt{bfloat16} precision with a learning rate of $2\times 10^{-6}$. Models are trained for 3 epochs, and each fine-tuning run requires approximately 48 hours.

\bmhead{Data availability}
Access to raw ADNI data requires user registration and adherence to the ADNI Data Use Agreement \footnote{ADNI DUA: \url{https://adni.loni.usc.edu/wp-content/themes/adni_2023/documents/ADNI_Data_Use_Agreement.pdf}}.
The authors do not have the right to redistribute the original ADNI data.
The processed data are available from the corresponding author upon reasonable request, subject to ADNI data-use regulations.

\bmhead{Code availability}
Code is available at \url{https://github.com/hegehongcha/genetic_imaging}
\bmhead{Acknowledgment}
This study is partially supported by the National Institutes of Health (R21AG087888, U01AG068057) and the National Science Foundation (CCF 2523787, IIS 2319450, IIS 2045848).
Part of this work used the Bridges-2 system, which is supported by NSF OAC-1928147 at the Pittsburgh Supercomputing Center (PSC).
We also acknowledge the UTRGV High Performance Computing Resource, supported by NSF grants 2018900 and IIS-2334389, and DoD grant W911NF2110169. 
Data used in preparation of this article were obtained from the Alzheimer’s Disease Neuroimaging Initiative (ADNI) database\footnote{\url{http://adni.loni.usc.edu}} funded by NIH grant U19AG024904. 
As such, the investigators within the ADNI contributed to the design and implementation of ADNI and/or provided data but did not participate in analysis or writing of this report. 
A complete listing of ADNI investigators can be found at: \footnote{\url{http://adni.loni.usc.edu/wp-content/uploads/how_to_apply/ADNI_Acknowledgement_List.pdf.}}

\bmhead{Author contributions}
K.Z. contributed to conceptualization, formal analysis, investigation, methodology, validation, visualization, writing—original draft, and writing—review \& editing. 
S.D., Y.Z., G.L., and P.G. contributed to conceptualization, methodology, and writing—review \& editing.  
C.L. contributed to methodology, formal analysis, validation and investigation.
P.M.T., A.L., H.H. and L.H. contributed to writing—review \& editing, funding acquisition and validation.  
L.Z. and H.T. contributed to data curation, formal analysis, funding acquisition, investigation, resources, supervision, writing—review \& editing and project administration.

\bmhead{Competing interests}
The authors declare no financial or non-financial competing interests related to this work.

\bibliography{reference, reference2}% common bib file

\end{document}